\documentclass{article}

\usepackage{microtype}
\usepackage{graphicx}
\usepackage{booktabs} 
\usepackage{mathtools}
\usepackage{caption}
\usepackage[labelformat=simple]{subcaption}
\usepackage{amsfonts}
\usepackage{amsmath,amsfonts,amssymb}
\newcommand{\R}{\mathbb{R}}
\usepackage[toc,page]{appendix}

\usepackage{hyperref}

\usepackage[accepted]{icml2020}

\icmltitlerunning{Learning to Branch for Multi-Task Learning}

\begin{document}

\twocolumn[
\icmltitle{Learning to Branch for Multi-Task Learning}



\icmlsetsymbol{equal}{*}

\begin{icmlauthorlist}
\icmlauthor{Pengsheng Guo}{to}
\hspace{5mm}
\icmlauthor{Chen-Yu Lee}{to}
\hspace{5mm}
\icmlauthor{Daniel Ulbricht}{to}

\end{icmlauthorlist}

\icmlaffiliation{to}{Apple}

\icmlcorrespondingauthor{Pengsheng Guo}{pengsheng\_guo@apple.com}

\icmlkeywords{Machine Learning, ICML, Multi-Task Learning, Neural Architecture Search}

\vskip 0.3in
]



\printAffiliationsAndNotice{}  

\begin{abstract}
Training multiple tasks jointly in one deep network yields reduced latency during inference and better performance over the single-task counterpart by sharing certain layers of a network. However, over-sharing a network could erroneously enforce over-generalization, causing negative knowledge transfer across tasks. Prior works rely on human intuition or pre-computed task relatedness scores for ad hoc branching structures. They provide sub-optimal end results and often require huge efforts for the trial-and-error process.

In this work, we present an automated multi-task learning algorithm that learns where to share or branch within a network, designing an effective network topology that is directly optimized for multiple objectives across tasks. Specifically, we propose a novel tree-structured design space that casts a tree branching operation as a gumbel-softmax sampling procedure. This enables differentiable network splitting that is end-to-end trainable. We validate the proposed method on controlled synthetic data, CelebA, and Taskonomy. 
\end{abstract}

\section{Introduction}
Multi-task learning~\cite{caruana1997multitask} has experienced rapid growth in recent years. Because of the breakthroughs in the performance of individually trained single-task neural networks, researchers have shifted their attention towards training networks that are able to solve multiple tasks at the same time.
One clear benefit of such a system is reduced latency where one network can produce multiple predictions in one forward propagation. This is particularly critical for portable devices that have limited computational budget.
Moreover, when training with various supervisory signals, it induces inductive bias ~\cite{mitchell1980need} where a network prefers some hypotheses over other hypotheses. From the point of view of any single task, the other tasks serve as regularizers in a sense that the network is asked to form representations that explain well for more than needed for solving one task, potentially improving generalization.

Contrary to the conventional single-task paradigm, training multiple tasks simultaneously in one network often encounters many challenges: some tasks are easier to train than others, some tasks have noisier ground truth labels than others, and some tasks are equipped with loss functions that have drastically different scales than others such as L1 vs cross-entropy.
Most of the work done in this field has focused on establishing some sort of parameter sharing mechanism by either sharing the whole network across all tasks or 
by assigning each task an individual set of parameters with crosstalk connections between tasks~\cite{ruder2017overview}. 
However, it is prohibitively expensive to design an optimal parameter sharing schema based on human intuition.
Another line of work has tried to balance the importance of different tasks by manipulating relative weighting between each task's loss ~\cite{kendall2018multi, guo2018dynamic, chen2017gradnorm}. But weight balancing alone also limits the potential performance gain under a fixed pre-defined network architecture.

There are many ways a network can invest its capacity for different tasks, and the design choice has a fundamental impact on its learning dynamics and final performance. Note that an exhaustive search of an optimal parameter sharing schema has combinatorial complexity as the number of tasks grows.
Prior literature has presented evidence that multi-task learning in back-propagation networks discovers task relatedness without the need of supervisory signals, and has presented results with k-nearest neighbor and kernel regression models \cite{caruana1997multitask}.
In this work we ask the following question: \textit{is it possible to automatically search a network topology based on the back-propagation signals computed from the multi-task objective?}

Typical neural networks learn the hierarchical nature of the feature representations. Specifically for computer vision applications, convolutional neural networks tend to learn more general feature representations in earlier layers such as edges, corners, and conjunctions ~\cite{zeiler2014visualizing}. We therefore expect a network at least shares the first few layers across tasks.
A key challenge towards answering the question is then deciding what layers should be shared across tasks and what layers should be untied. Over-sharing a network could erroneously enforce over-generalization, causing negative knowledge transfer across tasks.
In this work, we propose a tree-structured network design space that can automatically learn how to branch a network such that the overall multi-task loss is minimized. The branching operation is executed by sampling from a categorical latent variable formed by gumbel-softmax distribution ~\cite{jang2016categorical}.
This data-driven network structure searching approach does not require prior knowledge of the relationship between tasks nor human intuition on what layers capture task-specific features and should be split.

\section{Related Work}
Thanks to the genericness and the transferability of a number of off-the-shelf neural networks \cite{simonyan2014very, szegedy2015going, he2016deep} pre-trained on a large collection of samples, most of the prior works in the domain of multi-task learning are based on these popular backbone architectures and can be commonly categorized into either soft parameter sharing or hard parameter sharing \cite{ruder2017overview}.

In the soft sharing setting, each task has its own set of backbone parameters with some sort of regularization mechanisms to enforce the distance between weights of the model to be close.
Neural Network Parser ~\cite{duong2015low} uses two backbones, one for source language and one for target language, to perform multi-task learning. An extra set of weights are used for cross-lingual knowledge sharing by connecting activations between the source and target language model.
Cross-Stitch Networks ~\cite{misra2016cross} utilize an extra set of shared units to combine the activations between backbones from multiple tasks and learn the strength of information flow between tasks from data.
However, the total number of parameters in a soft parameter sharing system grows linearly with the number of tasks in general. Such approaches may encounter over-fitting due to lack of sufficient training samples to support the parameter space of multiple full-size backbones across all modalities or require higher computational cost during inference.

In the hard sharing setting, all tasks share the same set of backbone parameters, or at least share part of the backbone with branches toward the outputs. Deep Relationship Networks ~\cite{long2015learning} share the first five convolutional layers of AlexNet ~\cite{krizhevsky2012imagenet} among the tasks and use task-specific fully-connected layers tailored to fit different tasks.
Fully-Adaptive Feature Sharing method ~\cite{lu2017fully} starts with a thin network and expands it layer-by-layer based on the difficulty of the training set in a greedy fashion.
UberNet ~\cite{kokkinos2017ubernet} jointly solves seven labelling tasks by sharing an image pyramid architecture with tied weights. Meta Multi-Task Learning ~\cite{ruder2019latent} uses a shared input layer and two task-specific output layers.
Nonetheless, it is still unclear how to effectively decide what weights to share given a network  with a set of tasks in interest.

Instead of choosing between soft sharing or hard sharing approach, a new effort in tackling the multi-task learning problem is to consider the dynamics between different losses across tasks. Uncertainty-based weighting approach ~\cite{kendall2018multi} weighs multiple loss functions by utilizing the homoscedastic uncertainty of each task. GradNorm ~\cite{chen2017gradnorm} manipulates the magnitude of gradients from different loss functions to balance the learning speed between tasks. Task Prioritization ~\cite{guo2018dynamic} emphasizes more difficult tasks by adjusting the mixing weight of each task's loss objective. Multi-Objective Optimization approach ~\cite{sener2018multi} casts the multi-task learning problem as finding a set of solutions that lies on the Pareto optimal boundary. These methods can automatically tune the weightings between tasks and are especially effective when dealing with loss functions with different scales such as L1, L2, cross-entropy, etc. Yet, they use pre-defined network structures that might lead to sub-optimal solutions when the network topologies remain static.

The general focus in Neural Architecture Search (NAS) literature~\cite{zoph2016neural, zoph2018learning, liu2018progressive, wong2018transfer, liu2018darts, shaw2019meta, pasunuru2019continual} is on finding a repetitive cell or a global structure that is optimized over a single classification loss with a few exceptions that include memory or power constraint.
To better utilize the parameters of a network for multiple tasks, recently some works present methods to dynamically distribute the network capacity based on the complexities of the tasks and relatedness between the tasks.
Soft Layer Ordering ~\cite{meyerson2017beyond} learns to generate a task-specific scaling tensor to manipulate the magnitude of feature activations at different layers.
Evolutionary Architecture Search ~\cite{liang2018evolutionary} improves upon the Soft Layer Ordering by a synergetic approach of evolving custom shared routing units.
Soft attention used in ~\cite{liu2019end} allows learning of task-specific feature-level weighting for each task.
AdaShare ~\cite{sun2019adashare} learns the sharing pattern through a task-specific policy that selectively chooses which layers to execute for a given task in a multi-task setting.
The authors in ~\cite{standley2019tasks} propose to discover an optimal network splitting structure for different tasks by approximating the enumerative search process so that the test performances are maximized given a fixed testing resource budget.
Branched Multi-task Networks ~\cite{vandenhende2019branched} pre-compute a collection of task relatedness scores based on the usefulness of a set of features of one task for the other.
Gumbel-Matrix Routing ~\cite{maziarz2019gumbel} stacks a fixed set of operations in each layer and learns the connectivities between layers. These works do not rely on fixed network connectivities and start to explore the potential of more dynamic network wirings tailored to multiple tasks. 
But on the other hand, additional computation is often required for obtaining task relatedness scores in order to perform task grouping or splitting.

In this paper we propose a new end-to-end trainable algorithm that can automatically design a hard parameter sharing multi-task network, sharing and splitting network branches based on the update gradients back-propagated from the overall losses across all tasks. The proposed method bypasses the need of pre-computed task relatedness scores and directly optimizes over the end outputs, saving tedious computation and producing effective network topologies.

\vspace{-1mm}
\section{Method}
\vspace{-1mm}
We introduce the formal problem definition in Section ~\ref{fsetup}. We present the proposed network design space in Section ~\ref{tspace} and the differentiable branching operation formulation in Section ~\ref{dbranch}. Finally we show how the final network architecture is selected after training in Section ~\ref{fselect}.

\vspace{-1mm}
\subsection{Formulation Setup}
\vspace{-1mm}
\label{fsetup}
Given a set of $N$ tasks $\mathcal{T} = \{t_1, t_2, ..., t_{N}\}$, the goal of the proposed method is to learn a tree-structured~\cite{lee2016generalizing} network architecture $\Omega$ and the weight values $\omega$ of the network that minimize the overall loss $\mathcal{L}_{\text{total}}$ across all tasks,
\begin{equation}
\small
\begin{split}
\omega^*, \Omega^* 
& = \arg\min_{\omega, \Omega} \mathcal{L}_{\text{total}}(\omega, \Omega) \\
& = \arg\min_{\omega, \Omega} \sum_k \alpha_k \mathcal{L}_{k} (\omega, \Omega) \\
\end{split}
\end{equation}
where $\mathcal{L}_{k}$ is the loss for task $k$ and $\alpha_k$ is the task-specific weighting. The tree structure in the network is realized by branching operations at certain layers. Each branching layer can have an arbitrary number of child (next) layers up to the computational budget available.

During training, we first sample a network configuration $\Omega$ from the design space distribution and then perform forward propagation to compute the overall loss value $\mathcal{L}_{\text{total}}$. We then obtain corresponding gradients to update both the design space distribution and the weight matrices $\omega$ in the network in backward fashion. We iterate through the process until the overall validation loss converges and then we sample our final network configuration using the converged design space distribution.

\begin{figure}[t]
    \centering
    	\centering\includegraphics[width=0.3\textwidth]{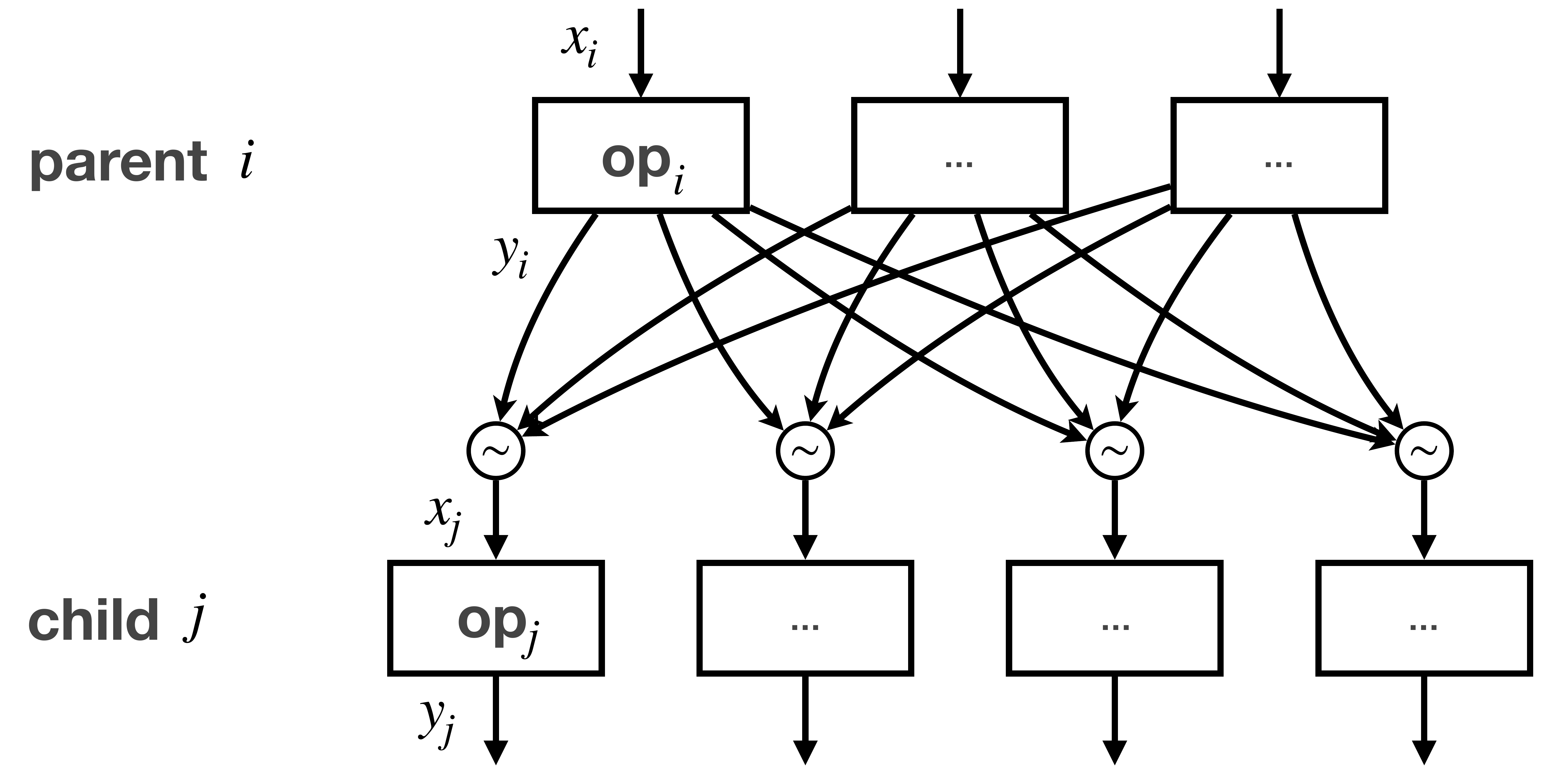}
    \vspace{-1mm}
  \caption{Illustration of the proposed branching block. Each child node $j$ is equipped with a categorical distribution so it can sample a parent node to receive input data after the training.}
  \label{topologyspace}
  \vspace{-2mm}
\end{figure}

\vspace{-1mm}
\subsection{Network Topological Space}
\vspace{-1mm}
\label{tspace}

The key ingredient for effective and efficient network configuration sampling is our proposed differentiable tree-structured network topology.
The topological space is represented as a Directed Acyclic Graph (DAG) where the nodes represent computational operations and the edges denote data flows.
Figure~\ref{topologyspace} illustrates a certain block of a DAG which contains parent nodes $i$ for $i \in \{1, ..., I\}$ and child nodes $j$ for $j \in \{1, ..., J\}$. The nodes can perform any common operations of choice such as convolution or pooling. The input to a certain node is denoted as $x$ and the output is denoted as $y$.

Specifically, we construct multiple parent nodes and child nodes for each block and allow a child node to sample a path from all the paths between it and all its parent nodes. The selected connectivities therefore define the tree structure by such sampling (branching) procedure. We formulate the branching operation at layer $l$ as:
\begin{equation}
\label{dotp}
\small
x^{l+1}_j = \mathbb{E}_{d_j \sim p_{\theta_j}} [\,d_j \cdot  Y^{l}\,]
\end{equation}
where $Y^{l} = [\,y_1^{l}, ..., y_I^{l}\,]$ concatenates outputs from all parent nodes at layer $l$, and $d_j$ is an indicator vector sampled from a certain distribution $p_{\theta_j}$. The indicator $d_j$ is a one-hot vector. Hence the dot product in Eq ~\ref{dotp} essentially assigns one of the parent nodes to each child node $j$. In other words, each parent node at layer $l$ propagates its output activations as input $x^{l+1}_j$ to one or more child nodes $j$ based on the sampling distributions. The sampling distribution is parameterized by $\theta_j$. The proposed topological space degenerates into a conventional single-path (convolutional) neural network if each block only contains one parent node and one child node.

\begin{figure*}[ht]
\begin{center}
  \begin{subfigure}[t]{.22\textwidth}
    	\centering\includegraphics[width=\textwidth]{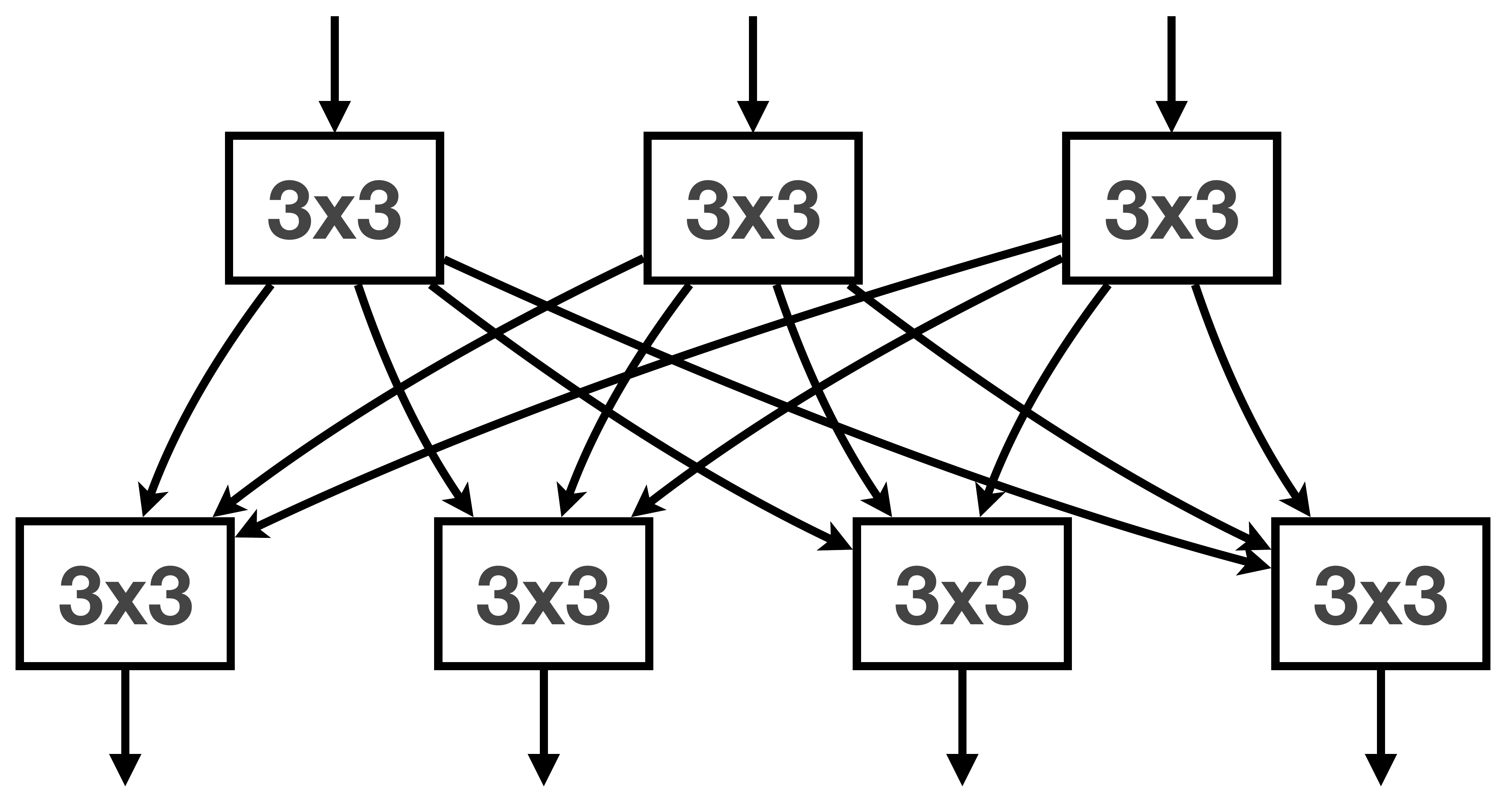}
	\vspace{-6mm}
    	\caption{}
  \end{subfigure}
  \hspace{3mm}
  \begin{subfigure}[t]{.22\textwidth}
    	\centering\includegraphics[width=\textwidth]{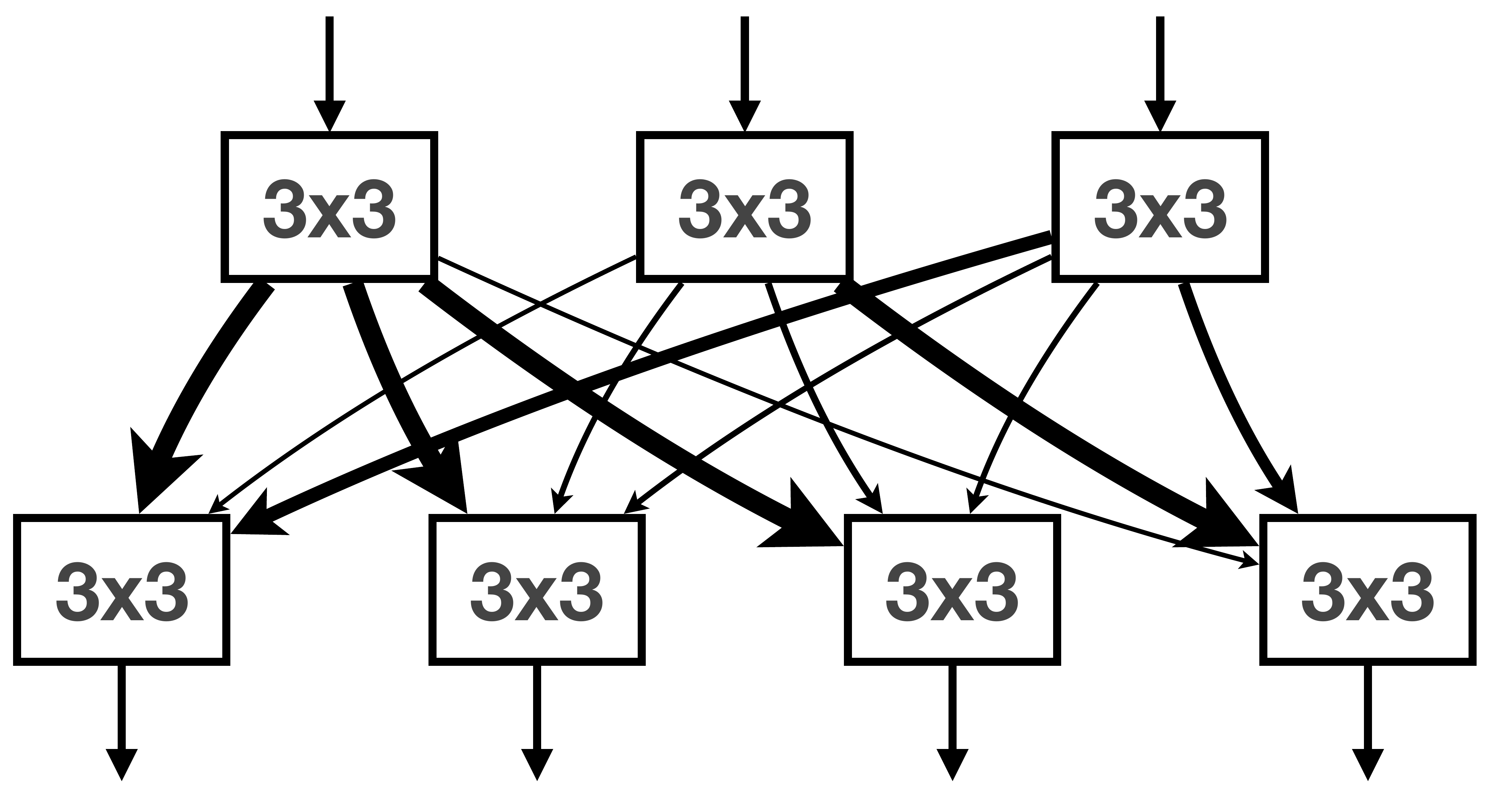}
	\vspace{-6mm}
    	\caption{}
  \end{subfigure}
  \hspace{3mm}
  \begin{subfigure}[t]{.22\textwidth}
    	\centering\includegraphics[width=\textwidth]{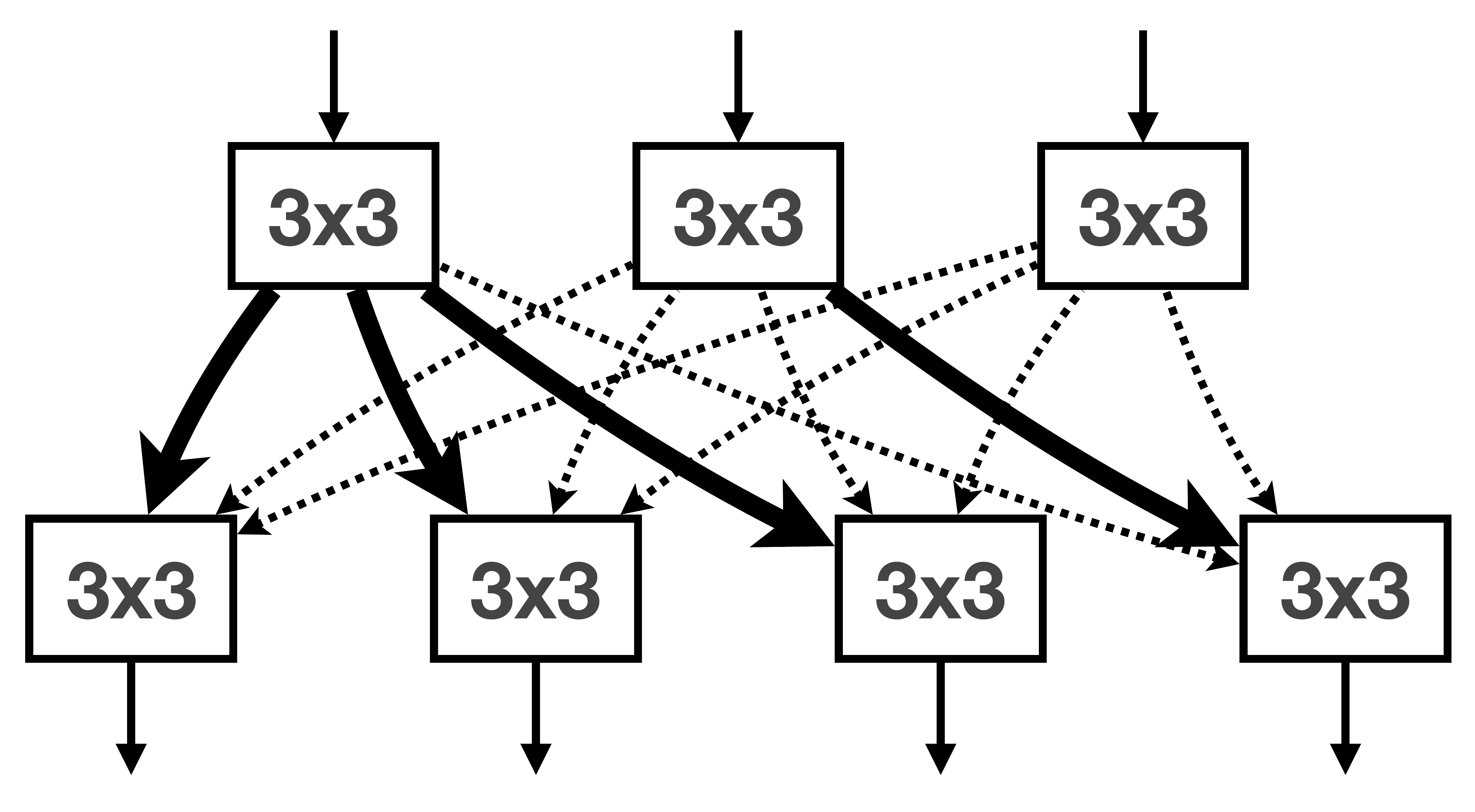}
	\vspace{-6mm}
    	\caption{}
  \end{subfigure}
  \hspace{3mm}
  \begin{subfigure}[t]{.22\textwidth}
    	\centering\includegraphics[width=\textwidth]{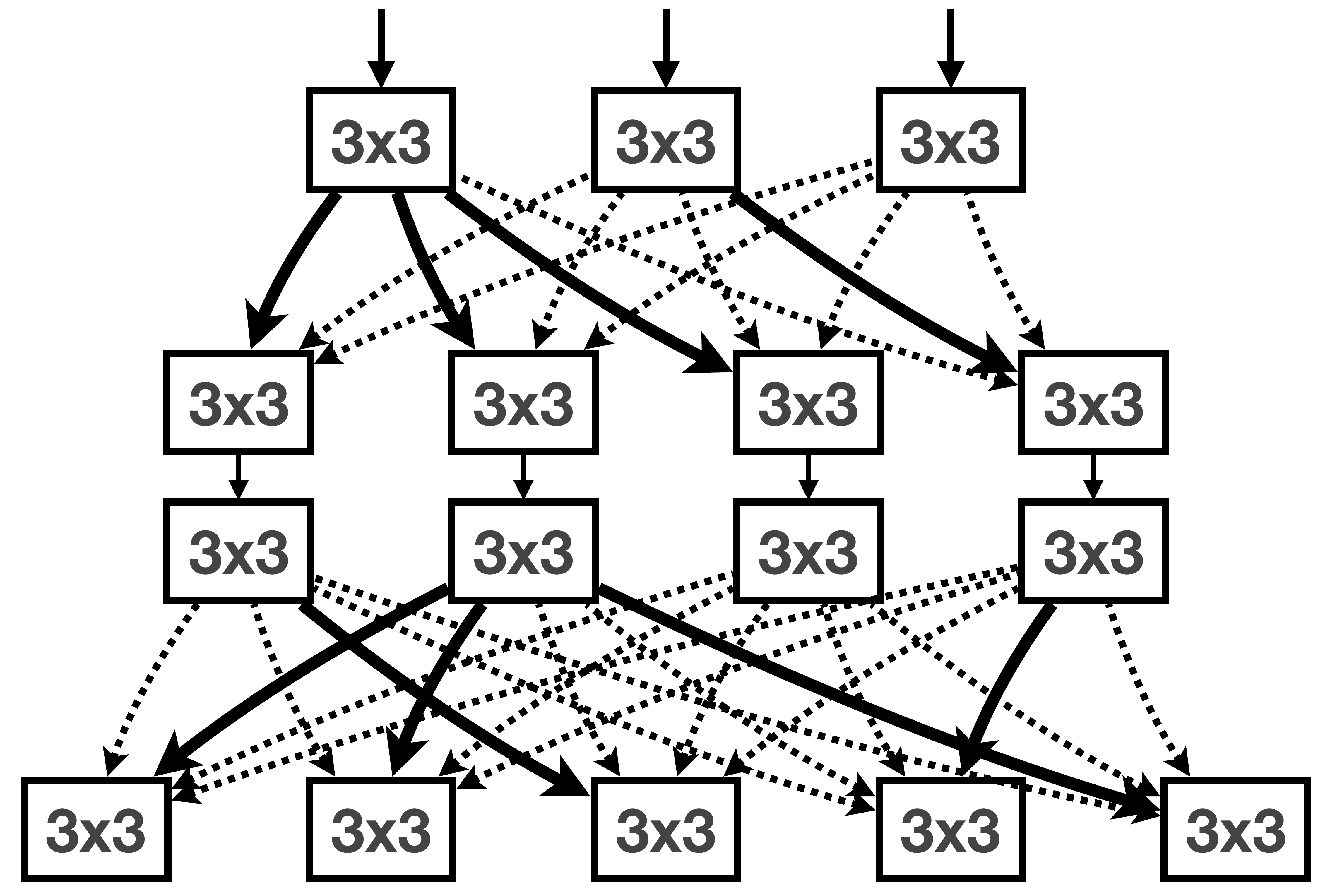}
	\vspace{-6mm}
    	\caption{}
  \end{subfigure}
  	\vspace{-3mm}
  \caption{Illustrations of the proposed learning to branch pipeline. 
(a) We initialize the sampling probability with a uniform distribution so each parent node has an equal chance to send its activation values to a child node.
(b) The computed update gradients then increase the probability of sampling certain paths that are more likely to reduce the overall loss.
(c) Once the overall validation loss converges, each child node selects one parent node with the highest sampling probability while removing unselected paths and parent nodes.
(d) We can construct a deeper tree-structured multi-task neural network by stacking such branching blocks.}
\label{s_p}
\vspace{-3mm}
\end{center}
\end{figure*}

We update the parameter $\theta_j$ of the sampling distribution $p_{\theta_j}$ using the chain rule with respect to the final loss,
\begin{equation}
\small
\begin{split}
\frac{\partial \mathcal{L}_{\text{total}}}{\partial {\theta_j}} 
& = \frac{\partial \mathcal{L}_{\text{total}}}{\partial x^{l+1}_j}  \frac{\partial x^{l+1}_j}{\partial {\theta_j}} \\
& = \frac{\partial \mathcal{L}_{\text{total}}}{\partial x^{l+1}_j}  \frac{\partial}{\partial {\theta_j}} \, \mathbb{E}_{d_j \sim p_{\theta_j}} [\,d_j \cdot  Y^{l}\,] \\
\end{split}
\label{eq:dist_g}
\end{equation}
the backward pass then adjusts the sampling distribution $p_{\theta_j}$ to make it more likely to generate network configurations $\Omega$ toward the direction of minimizing the overall loss $\mathcal{L}_{\text{total}}$.

The branching blocks in Figure~\ref{topologyspace} can be stacked to form a deeper tree-structured neural network (illustrated in Figure~\ref{s_p}(d)) and the number of parent nodes and the number of child nodes can be adjusted based on the desired model capacity.
Different from the greedy layer-wise optimization approach in GNAS~\cite{huang2018gnas}, our proposed tree-structured network topology is end-to-end trainable -- the network architecture $\Omega$ and the weight matrices $\omega$ of the network are jointly optimized during training.

\subsection{Differentiable Branching Operation}
\label{dbranch}
To sample a categorical value from the continuous sampling distribution, we utilize the gumbel-softmax estimator trick~\cite{jang2016categorical, shazeer2017outrageously, rosenbaum2017routing, veit2018convolutional, xie2018snas} to enable the differentiability for the branching operation; During the feedforward pass, we sample a parent node by a discrete index value based on a certain probability distribution for each child node. During the backward pass, we update the probability distribution by replacing the discrete samples with gumbel-softmax samples.

For every two layers in a branching block (shown in Figure~\ref{topologyspace}), we construct a matrix $M \in \R^{I \times J}$ to represent the connectivity from parent nodes $i$ to child nodes $j$. Each entry $\theta_{i, j}$ in such a matrix $M$ stores the probability value that represents how likely the parent node $i$ would be sampled to connect with the child node $j$. During every forward propagation, each child node $j$ makes a discrete decision drawn from a categorical distribution based on the distribution:
\begin{equation}
\small
d_j= \textrm{one\_hot} \{\arg\max_i (\log \theta_{i, j} + \epsilon_i) \}
\end{equation}
Again $d_j \in \R^{I}$ is a one-hot vector with dimension the same as the number of parent nodes $I$ at the current level. $\epsilon \in \R^{I}$ is a vector with i.i.d samples draw from gumbel distribution $(0,1)$ to add a small amount of noise to avoid the $\arg\max$ operation always selecting the element with the highest probability value.

To enable differentiability of the discrete sampling function, we use the gumbel-softmax trick~\cite{jang2016categorical} to relax $d_j$ during backward propagation as
\begin{equation}
\small
\tilde{d_{j}} = \frac{\textrm{exp}(( \log \theta_{i, j} + \epsilon_i ) / \tau)}{\sum_k \textrm{exp}((\log \theta_{k, j} + \epsilon_k )/\tau)} \\
\end{equation}
with $i$ equal to the sampled index value of parent node during forward pass. The discrete categorical sampling function is approximated by a softmax operation over the parent nodes, and the parameter $\tau$ is the temperature that controls how sharp the distribution is after the approximation.

We can now utilize the reparameterization trick for random sample $d_j$ and rewrite the Eq \ref{eq:dist_g} as
\begin{equation}
\small
\begin{split}
\frac{\partial \mathcal{L}_{\text{total}}}{\partial {\theta_j}} 
& = \frac{\partial \mathcal{L}_{\text{total}}}{\partial x^{l+1}_j}  \frac{\partial}{\partial {\theta_j}} \, \mathbb{E}_{\epsilon} [\, \tilde{d_{j}} \cdot Y^{l} \, ] \\
& = \frac{\partial \mathcal{L}_{\text{total}}}{\partial x^{l+1}_j}  \, \mathbb{E}_{\epsilon}  [ \,  \frac{\partial \tilde{d_{j}}} {\partial \theta_i} \, ] \, Y^{l} \\
\end{split}
\end{equation}
At this stage, the branching probabilities are fully differentiable with respect to the training loss and can readily be inserted to a neural network and stacked to construct a tree-structured neural network. We decay the temperature $\tau$ gradually during training so the network can explore freely in the early stage and exploit the converged topology distribution in the later stage.

\begin{figure*}[h]
    \centering
    \
    \vspace{2mm}
  \begin{subfigure}{.21\textwidth}
    	\centering\includegraphics[width=\textwidth]{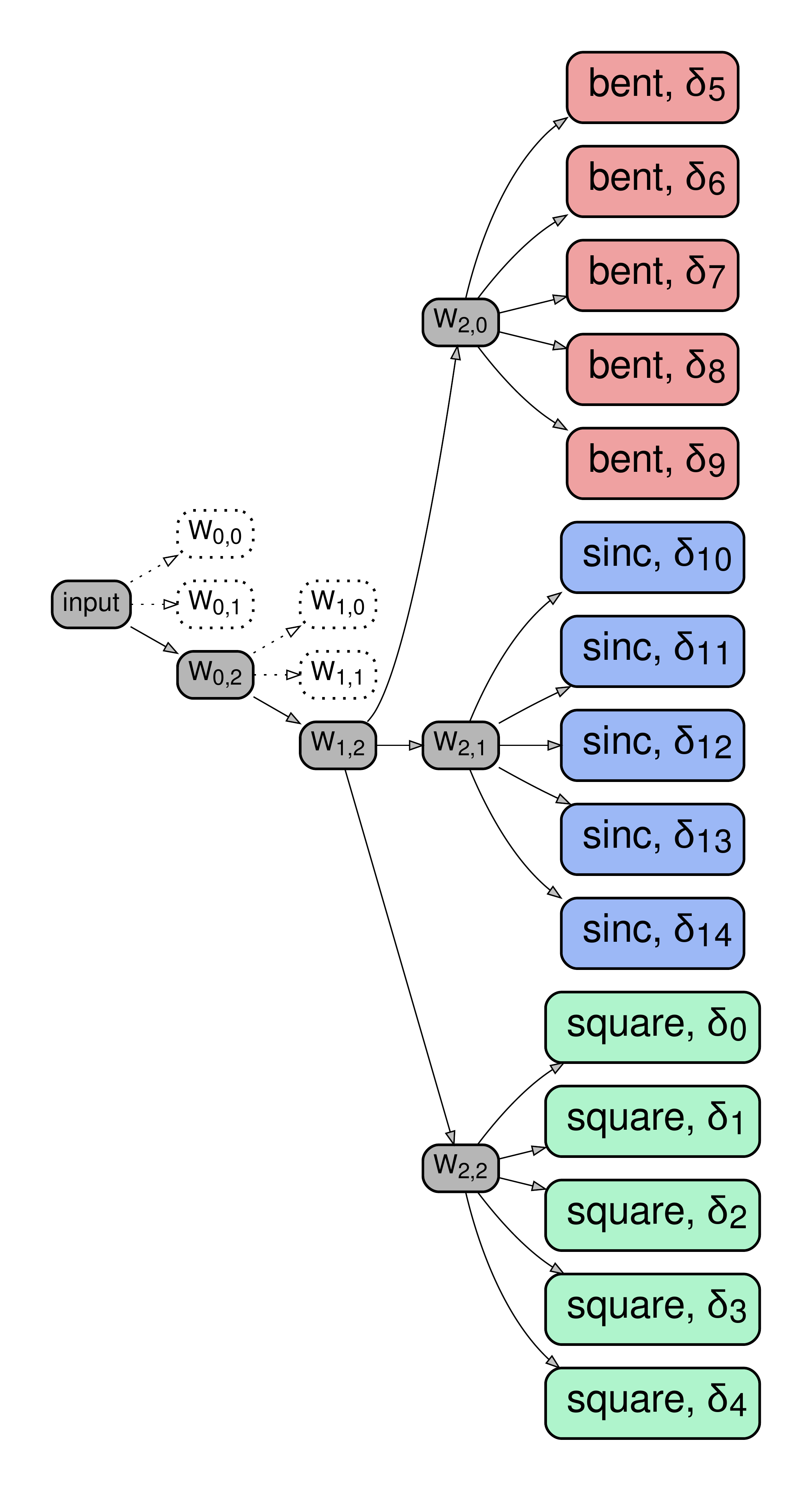}
	\vspace{-7mm}
    	\caption{}
    	\label{square_bent_sinc}
  \end{subfigure}
  \hspace{3mm}
  \begin{subfigure}{.21\textwidth}
      	\centering\includegraphics[width=\textwidth]{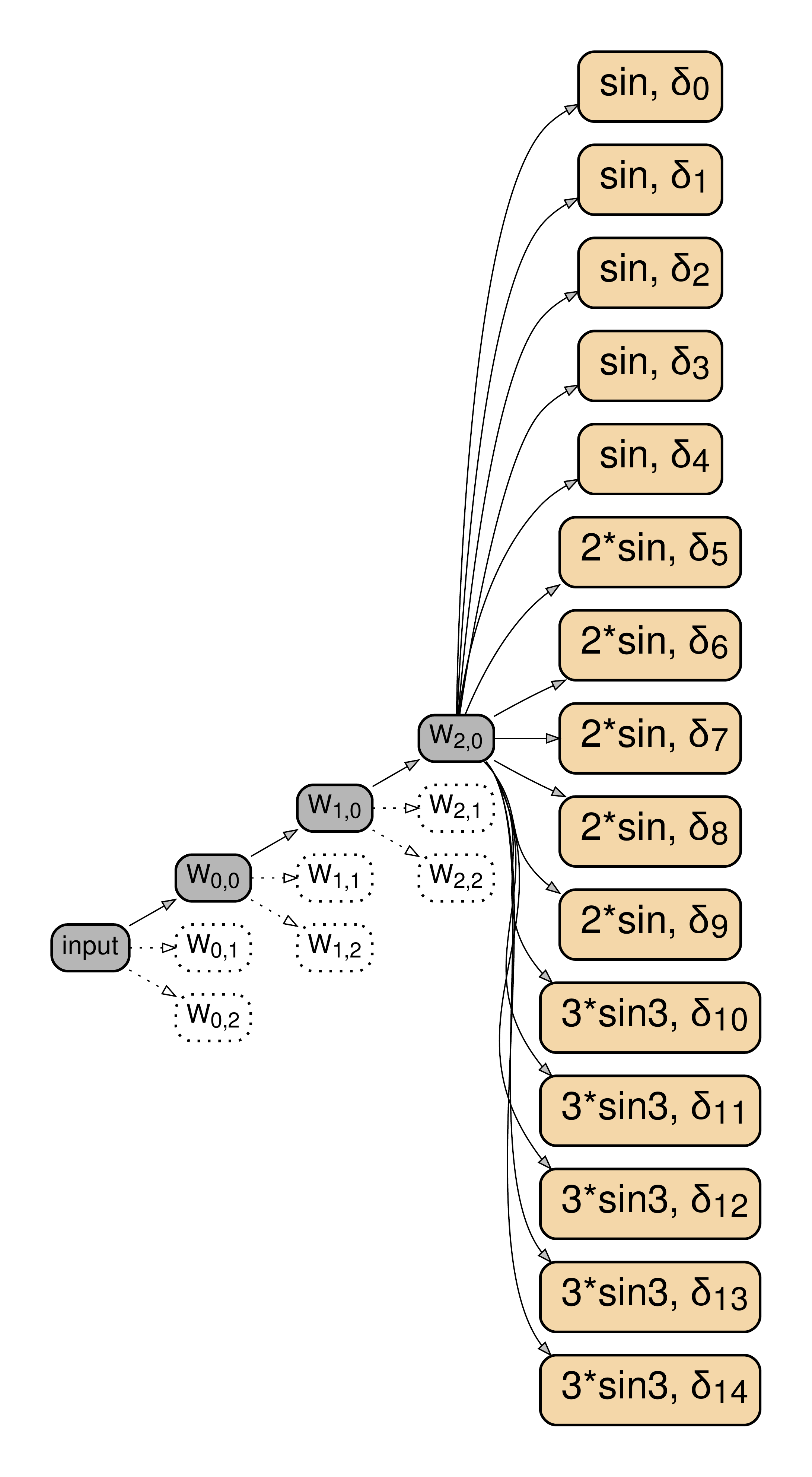}
	\vspace{-7mm}
    	\caption{}
	\label{sin_sin2_sin3}
  \end{subfigure}
 \hspace{3mm}
  \begin{subfigure}{.21\textwidth}
    	\centering\includegraphics[width=\textwidth]{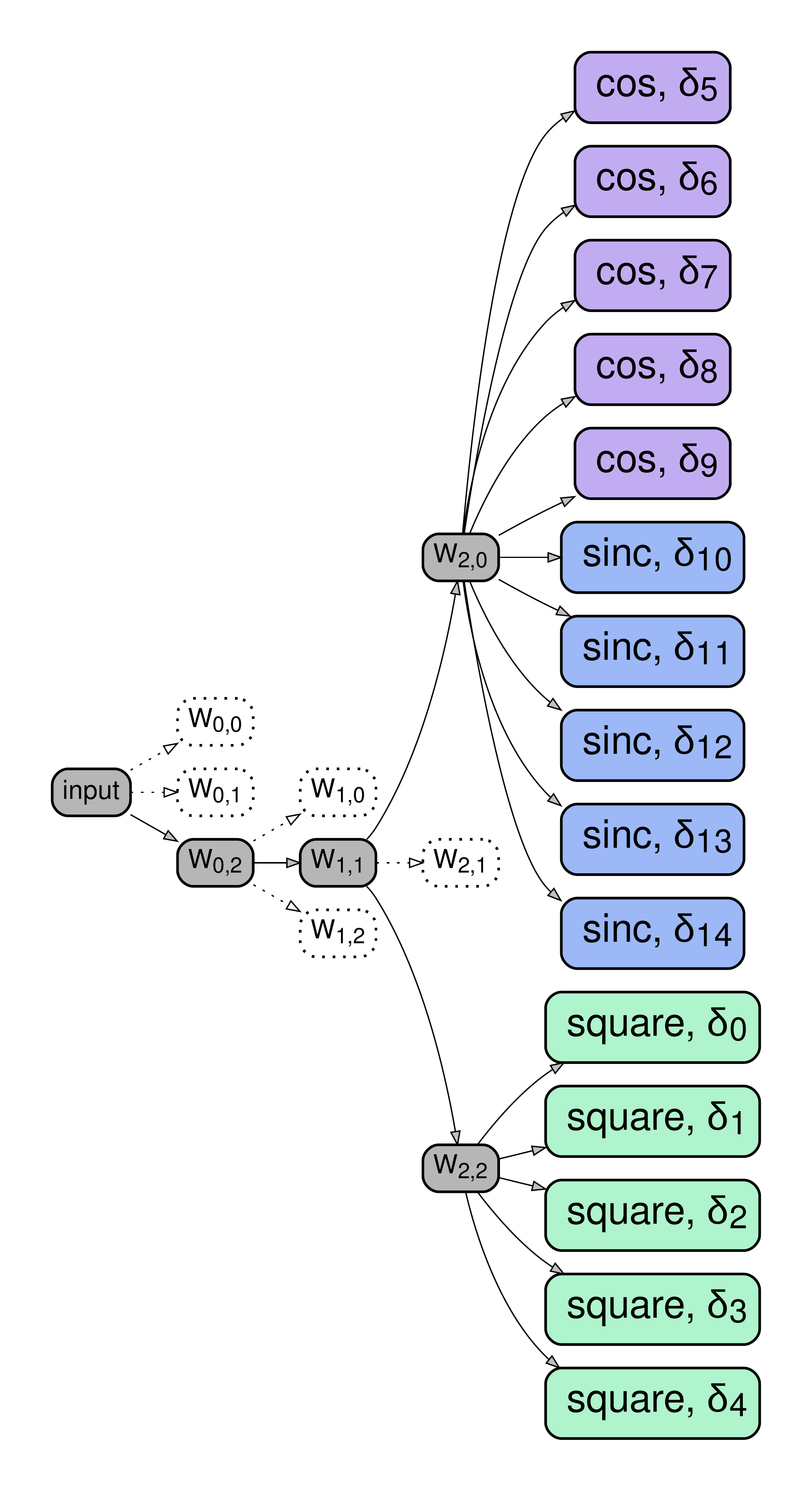}
	\vspace{-7mm}
    	\caption{}
	\label{square_cos_sinc}
  \end{subfigure}
 \hspace{15mm}
  \begin{subfigure}{.132\textwidth}
 	\vspace{-1mm}
    	\includegraphics[width=\textwidth]{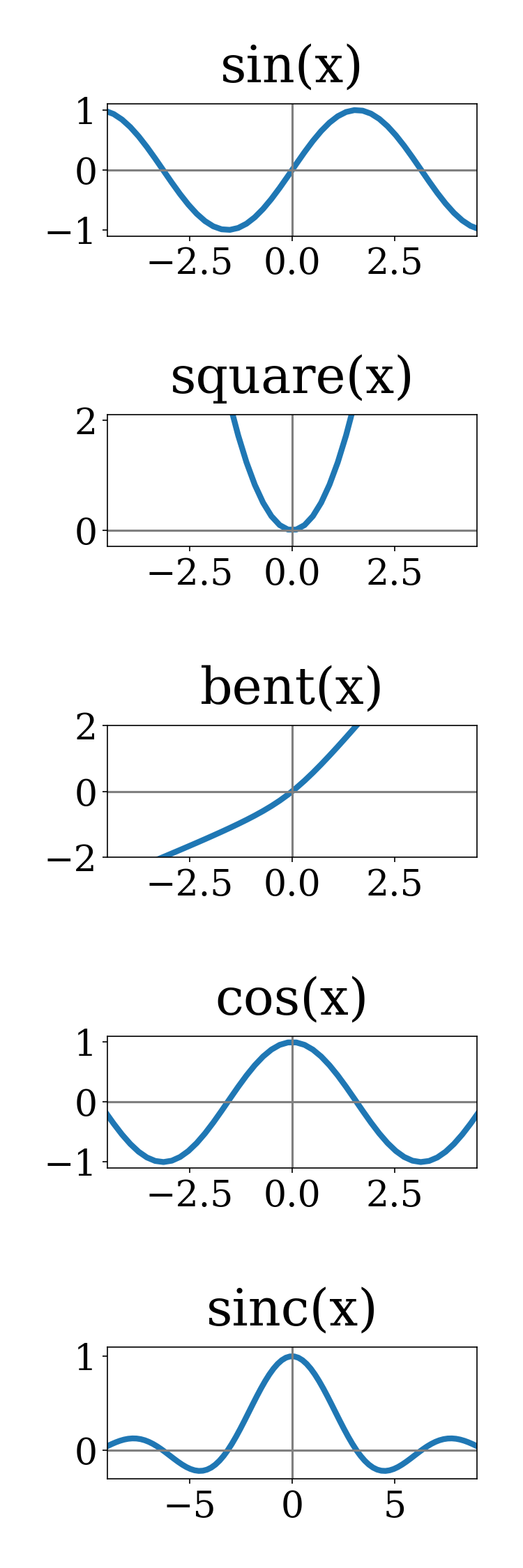}
   	 \vspace{-7mm}
	 \caption{}
	\label{act_visualization}
  \end{subfigure}
   \vspace{-4mm}
  \caption{Learned network architectures by our method in three different experimental settings -- each setting has $15$ tasks generated by $3$ different activation functions. 
Setting (a) contains activations bent, square, and sinc. The learned network group tasks with the same activation together in 3 distinct clusters.
Setting (b) contains the same activation sin but with different scale multipliers $1$, $2$, and $3$. All tasks are grouped together and share all intermediate layers.
Setting (c) contains activations cos, sinc, and square. As illustrated in (d) that cos and sinc  share similar behavior, tasks with these two similar activations are grouped while the task with square activation branches out earlier in the intermediate layer.}
    \vspace{-2mm}
       \label{syn}
\end{figure*}

\subsection{Final Architecture Selection}
\label{fselect}
During the training stage, the network topology distribution and the weight matrices of the network are jointly optimized over the loss $\mathcal{L}_{\text{total}}$ across all tasks. Once the validation loss converges, we simply select the final network configuration using the same categorical distribution but without the noise $\epsilon$ for every block in the network,
\begin{equation}
\small
d_j = \textrm{one\_hot} \{\arg\max_i ( \log \theta_{i, j})\}
\end{equation}
We then re-train the final network architecture from scratch to obtain the final performance. The same procedure has also been shown effective in previous literature~\cite{pham2018efficient, sciuto2019evaluating} where such weight sharing network search schema demonstrates high correlation between the intermediate network performance during search phase and the final performance obtained by re-train the network from scratch.

Figure \ref{s_p} illustrates our overall training process of a certain branching block in a DAG. We initialize the sampling probability $p_{\theta_j}$ with a uniform distribution so each parent node has an equal chance to send its activation values to a child node $j$ as shown in Figure \ref{s_p}(a).
The computed update gradients then increase the probability of sampling certain paths that are more likely to reduce the overall loss as shown in Figure \ref{s_p}(b).
Once the overall validation loss converges, each child node selects one parent node with the highest sampling probability while removing unselected paths and parent nodes as shown in Figure  \ref{s_p}(c).
We can construct a deeper tree-structured multi-task neural network by stacking such branching blocks as shown in Figure  \ref{s_p}(d) as long as we ensure the number of child nodes matches the number of tasks being trained. This process implicitly groups the relevant tasks together by sharing necessary layers. It does not require prior knowledge of the relatedness of the tasks and avoids exhausting trial-and-error searching process by hand.

\vspace{-1mm}
\section{Experiments}
\vspace{-1mm}
In principle, our method can be applied to any domains and does not require any prior knowledge of the tasks. We demonstrate the effectiveness of the proposed method on synthetic data for regression tasks, CelebA~\cite{liu2015faceattributes} dataset for classification tasks, and Taskonomy dataset~\cite{zamir2018taskonomy} for dense prediction that includes both regression and classification tasks.

\vspace{-1mm}
\subsection{Controlled Synthetic Data}
\vspace{-1mm}
We first validate the proposed concept using synthetic data with controllable task relatedness. The relatedness is realized by different activation functions. Inspired by \cite{chen2017gradnorm}, we construct the regression tasks by the formulation
\begin{equation}
\small
T_{r, s} = \text{activation}_r [(( B + \delta_s ) \, Z ) / \varphi ]
\end{equation}
where $r \in \{1, 2,  ..., 5\}$ corresponds to a range of five different element-wise activation functions  \{sin, square, bent, cos, sinc\}, $\delta_s $ denotes task-specific random noise matrices for $s \in \{1, 2, ..., S\}$. The input of the formulation $Z \in \R^{200}$ is a 200-dimensional vector and the output $T \in \R^{100}$ is a 100-dimensional vector. $B \in \R^{100 \times 200}$ and $\delta_s \in \R^{100 \times 200}$ are constant matrices randomly sampled from $\mathcal{N}(0, 10)$ and $\mathcal{N}(0, 2)$, respectively. $\varphi \in \R$ is a normalization term that has the value of size of the input dimension. 
We control the relatedness of the tasks by the activation used. Tasks constructed by the same activation function should be more related as they only differ in a small amount of random noise $\delta_s$. On the other hand, tasks with different activation functions should be more unrelated.

We use four proposed branching blocks to construct our tree-structured network for the multi-task learning setup. Each block has three child nodes with fully-connected layers as the choice of operation. Each fully-connected layer contains 100 neurons. We use simple bias terms for task-specific layers as shown in Figure~\ref{syn}. L2 loss is used as the training objective and the loss is optimized by the Adam solver with mini-batch size of $100$. Learning rate is set to $10^{-3}$ for weight matrices and $10^{-7}$ for branching probability throughout the training. Temperature is set to $50$ and decayed by the square root of the number of iterations. The networks are trained for $500$ epochs with $50$ epochs for warmup. We do not update the branching probability during warmup to ensure all weight matrices receive equal amounts of update gradients.

We perform experiments in three different settings -- each setting has equal weighted $15$ tasks generated by $3$ different activation functions.
In the first setting (Figure~\ref{square_bent_sinc}), we use activations bent, square, and sinc. The final learned network architecture clearly shows a tree structure with similar tasks (same activation) are grouped in the same leaf branch and dissimilar tasks (different activations) do not share the same leaf branch.
In the second setting (Figure~\ref{sin_sin2_sin3}), we use the same activation sin but with different scale multipliers $1$, $2$, and $3$. All tasks are grouped together and share all intermediate layers since they only differ in different scaling.
In the third setting (Figure~\ref{square_cos_sinc}), we use activations cos, sinc, and square. As illustrated in Figure~\ref{act_visualization} that cos and sinc activations share very similar active regions and scales, we can see that tasks with these two similar activations are grouped while the task with square activation branches out earlier in the intermediate layer. From this experiment, we validate our intuition that the proposed branching structure indeed captures the underlying task relatedness and is able to group related tasks through back-propagation updates.

\subsection{CelebA}

Next, we evaluate the proposed method on real-world image classification tasks. We use the CelebA dataset \cite{liu2015faceattributes}, which contains over $200$K face images and each image contains $40$ binary attribute annotations. Each annotation is regarded as a classification task and we adopt $40$ cross-entropy losses with equal weightings for all $40$ tasks.
The training, validation, and test sets contain 160K, 20K, and 20K images. 
This benchmark is especially useful to examine whether automatically learned task grouping is more effective than manual task grouping by human intuition or pre-computed task relatedness.

\textbf{Implementation details.}
For a fair comparison, we utilize the same overall network structures and operations from~\cite{lu2017fully, huang2018gnas} in our branching blocks but allow the network to learn the branching decisions.
Specifically, we construct (a) LearnToBranch-VGG model based on the Branch-VGG model in~\cite{lu2017fully} where the backbone is a truncated VGG19 \cite{simonyan2014very} with the number of channels reduced to 32 for  convolutional layers and 64 for the fully-connected layers, and (b) LearnToBranch-Deep-Wide based on the GNAS-Deep-Wide model in~\cite{huang2018gnas}, which is a customized architecture with 5 consecutive convolutional layers and 2 fully-connected layers. 
For model (a) we allow our network to branch at the end of each resolution stage (last conv layer at each resolution) with the number of child nodes set to $\{3, 3, 5, 5, 10, 20, 30, 40\}$, and for model (b) we allow our network to branch at the end of each convolutional layer with number of child nodes set to $\{2, 4, 8, 16, 40\}$. The number of child nodes are chosen so that the overall computational complexity of models (a) and (b) are similar to their counterparts.

We use the Adam optimizers with mini-batch size $64$ to update both the weight matrices and the branching probabilities in our networks. Temperature is set to $10$ and decayed by the number of epochs. We warmup the training for 2 epochs without updating the branching probabilities to ensure all weight matrices receive equal amounts of update gradients initially. Weight decay is set to $10^{-4}$ for all experiments. We perform grid search for learning rates in $(10^{-6}, 10^{-5}, 10^{-4})$ for the weights and in $(1, 10, 100)$ for  branching distributions. After sampling the final architecture, we train the network from scratch with grid search for global learning rate in $(0.02, 0.03, 0.04, 0.05)$. The input data is normalized $[-1, 1]$ and augmented by random flipping. Please refer to Appendix for more details.

\begin{table}[t]
\vspace{-1mm}
\caption{Results of multi-task learning on CelebA Dataset.}
\vspace{2mm}
\label{celeba-results}
\begin{center}
\begin{scriptsize}
\begin{sc}
\begin{tabular}{lcccr}
\toprule
Method & Acc (\%) & Params (M) \\
\midrule
LNet+ANet \cite{wang2016walk}    & 87 & -\\
Walk and Learn \cite{wang2016walk} & 88 & -\\
Moon  \cite{rudd2016moon}  &90.94 & 119.73\\
Indep Group  \cite{hand2017attributes}  & 91.06 & - \\
MCNN-AUX   \cite{hand2017attributes}   & 91.29 & - \\
VGG-16 Baseline  \cite{lu2017fully}    & 91.44 & 134.41 \\
\midrule
Branch-VGG   \cite{lu2017fully}   & 90.79 & 2.09 \\
\textbf{LearnToBranch-VGG (ours)}      & \textbf{91.55} & \textbf{1.94} \\
\midrule
GNAS-Deep-Wide   \cite{huang2018gnas}    & 91.36 & 6.41 \\
\textbf{LearnToBranch-Deep-Wide (ours)}      & \textbf{91.62} & \textbf{6.33} \\
\bottomrule
\end{tabular}
\end{sc}
\end{scriptsize}
\end{center}
\vspace{-4mm}
\end{table}

\begin{table*}[ht]
\caption{Results of multi-task learning on Taskonomy test set. Our method outperforms the direct comparable method AdaShare~\cite{sun2019adashare} and other baselines as well. Besides having fewer parameters and better performance, our method has a clear advantage of being the first end-to-end trainable tree-structured multi-task network that does not require human intuition or pre-computed task relatedness.}
\vspace{1mm}
\label{taskonomy-results}
\begin{center}
\begin{scriptsize}
\begin{sc}
\begin{tabular}{l |c | ccccccc}
\toprule
Method & Params (M) & Segmentation $\downarrow$ & Normal $\uparrow$ & Depth $\downarrow$ & Keypoint $\downarrow$ & Edge $\downarrow$\\
\midrule
Single-Task \cite{sun2019adashare}  & 124  & 0.575 & 0.707 & 0.022 & 0.197 & 0.212 \\
Multi-Task \cite{sun2019adashare}  &  41 & 0.587 & 0.702 & 0.024 & 0.194 & 0.201 \\
Cross-Stitch \cite{misra2016cross}  & 124   & 0.560 & 0.684 & 0.022 & 0.202 & 0.219 \\
Sluice \cite{Ruder2017SluiceNL}  & 124  & 0.610& 0.702 & 0.023 & 0.192& 0.198\\
NDDR-CNN  \cite{gao2019nddr}   &  133 & 0.539 &0.705 &0.024 & 0.194& 0.206\\
MTAN   \cite{liu2019end}   & 114  &0.637 &0.702 &0.023 & 0.193& 0.203\\
AdaShare \cite{sun2019adashare}    & 41 & 0.566& 0.707& 0.025& 0.192& 0.193\\
\midrule
\textbf{LearnToBranch (ours)}    & 51  & \textbf{0.462} & \textbf{0.709} & \textbf{0.018} & \textbf{0.122} & \textbf{0.136} \\
\midrule
\bottomrule
\end{tabular}
\end{sc}
\end{scriptsize}
\end{center}
\end{table*}

Our method leverages the effectiveness of gumbel-softmax so that every child node samples a single discrete action during the forward pass. Therefore, our network topological space is well maintained -- the tree does not grow exponentially with the number of tasks. 
As the result, it takes 10 hours to search the architecture and 11 hours to obtain the optimal weights for model (a) LearnToBranch-VGG, and it takes 4 hours to search the architecture and 10 hours to obtain the optimal weights for model (b) LearnToBranch-Deep-Wide on a single 16GB Tesla GPU.

\textbf{Results.}
Table~\ref{celeba-results} shows the performance comparison on the CelebA test set. The visualizations of the learned network architectures are provided in Appendix.
We can clearly see that both models (a) LearnToBranch-VGG and (b) LearnToBranch-Deep-Wide outperform their counterpart baselines presented in~\cite{lu2017fully, huang2018gnas} under the similar network capacity. In fact, both our models (a) and (b) have less total number of parameters than their baselines.
Note that our models only differ in the branching operation while maintaining other configurations such as kernel size and the number of channels. This demonstrates the effectiveness of the proposed end-to-end trainable branching mechanism.
We note that the ResNet-18 (MGDA-UB) model in~\cite{sener2018multi} achieves $91.75\%$ accuracy on this specific task. However their network has more than $11$ million parameters, which is roughly double the size of our model (b) with comparable performance. Also their focus is on reformulating the multi-task learning problem as multi-objective optimization. We propose to further study the possibility of combining the two techniques in future investigation.

\subsection{Taskonomy}
In this experiment we extend our method to the recent Taskonomy dataset~\cite{zamir2018taskonomy}, which contains over $4.5$ million indoor images from over $5,000$ buildings. Following~\cite{sun2019adashare}, we select surface normal, edge detection, keypoint detection, monocular depth, and semantic segmentation among the total $26$ tasks for the experiment. We use the standard tiny split benchmark, which contains $275$K training, $54$K test, $52$K validation images. We again follow the work in~\cite{sun2019adashare} to report test losses on these tasks for standardized comparisons.

\begin{figure*}[ht!]
    \centering
  \begin{subfigure}[t]{.42\textwidth}
    	\centering\includegraphics[width=\textwidth]{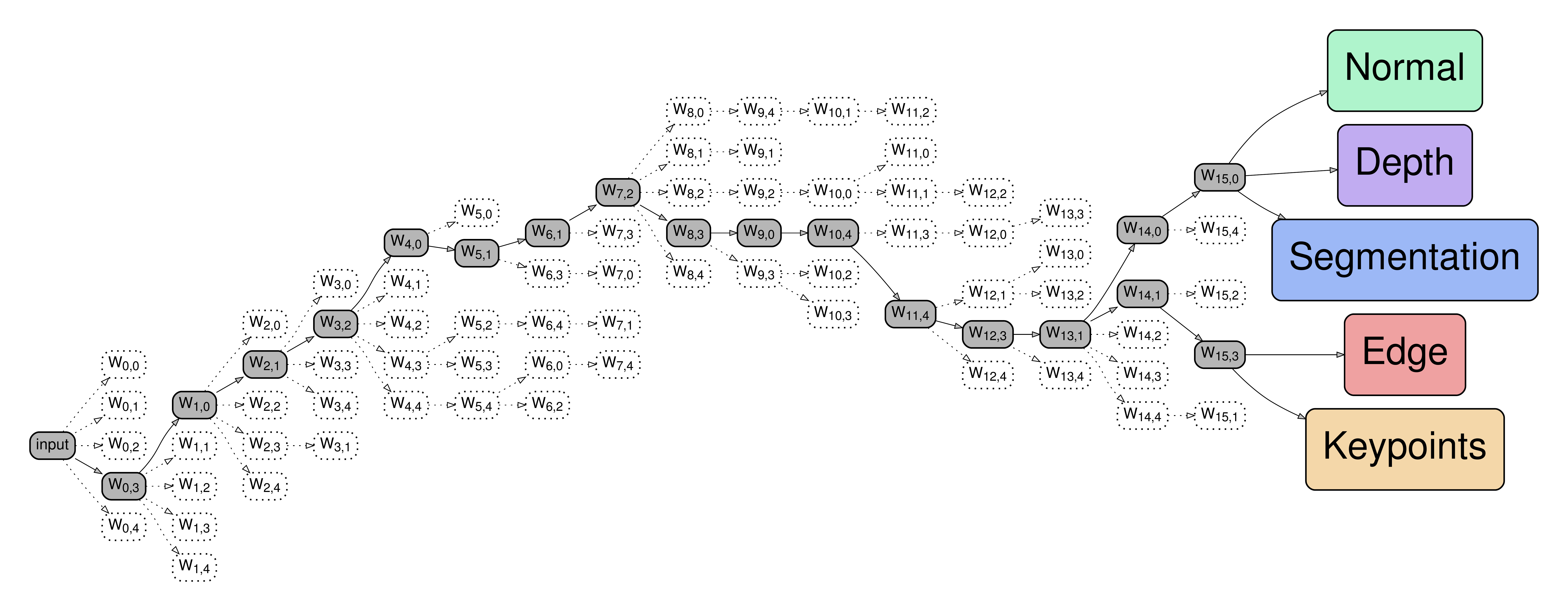}
	\vspace{-9mm}
    	\caption{}
	\label{taskonomy_1}
  \end{subfigure}
  \hspace{2mm}
  \begin{subfigure}[t]{.42\textwidth}
      	\centering\includegraphics[width=\textwidth]{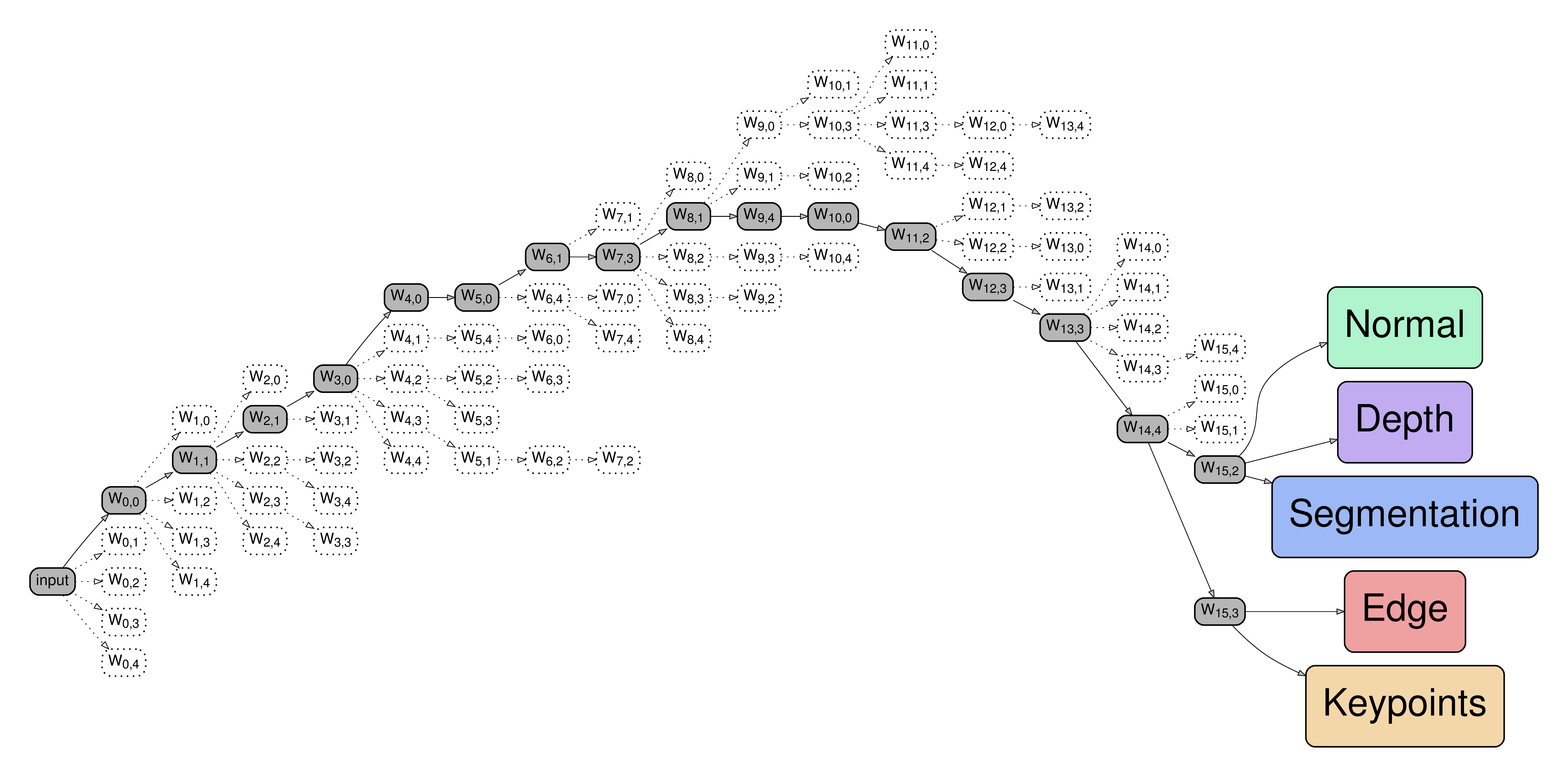}
	\vspace{-9mm}
    	\caption{}
	\label{taskonomy_2}
  \end{subfigure}
  
  \begin{subfigure}[t]{.42\textwidth}
    	\centering\includegraphics[width=\textwidth]{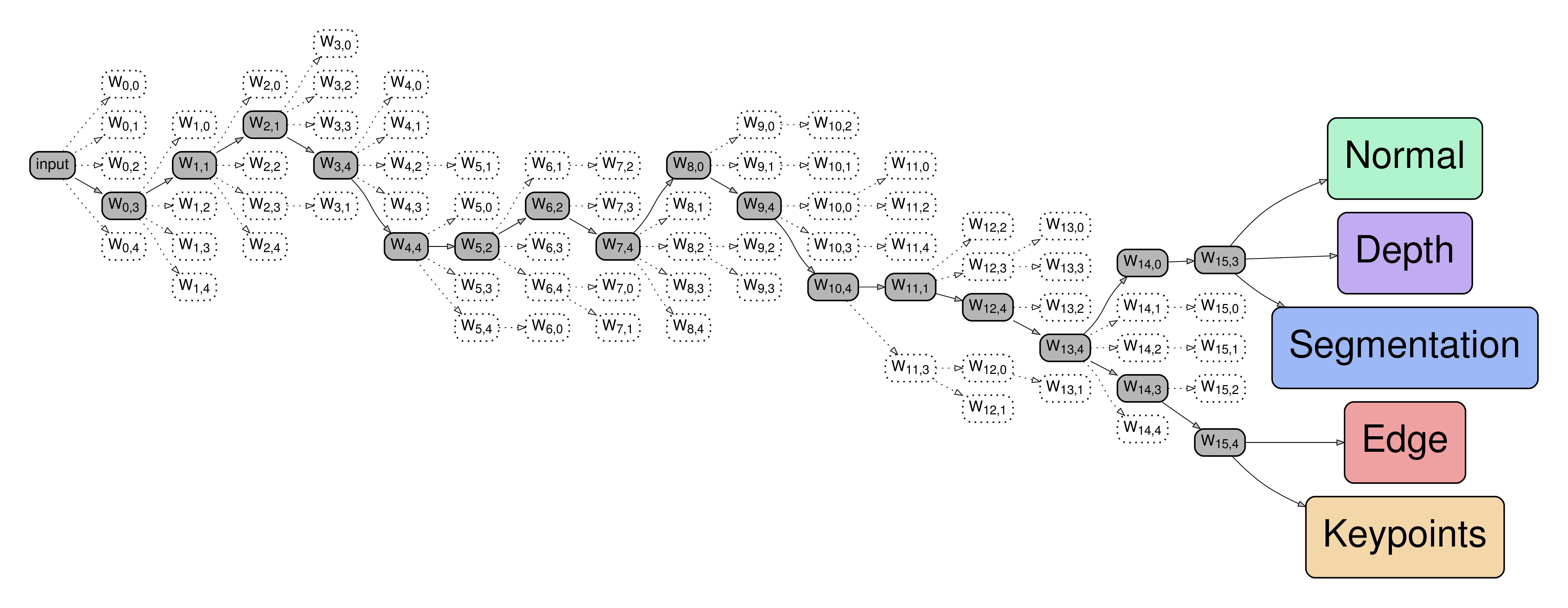}
	\vspace{-9mm}
    	\caption{}
	\label{taskonomy_3}
  \end{subfigure}
   \hspace{2mm}
  \begin{subfigure}[t]{.42\textwidth}
    	\centering\includegraphics[width=\textwidth]{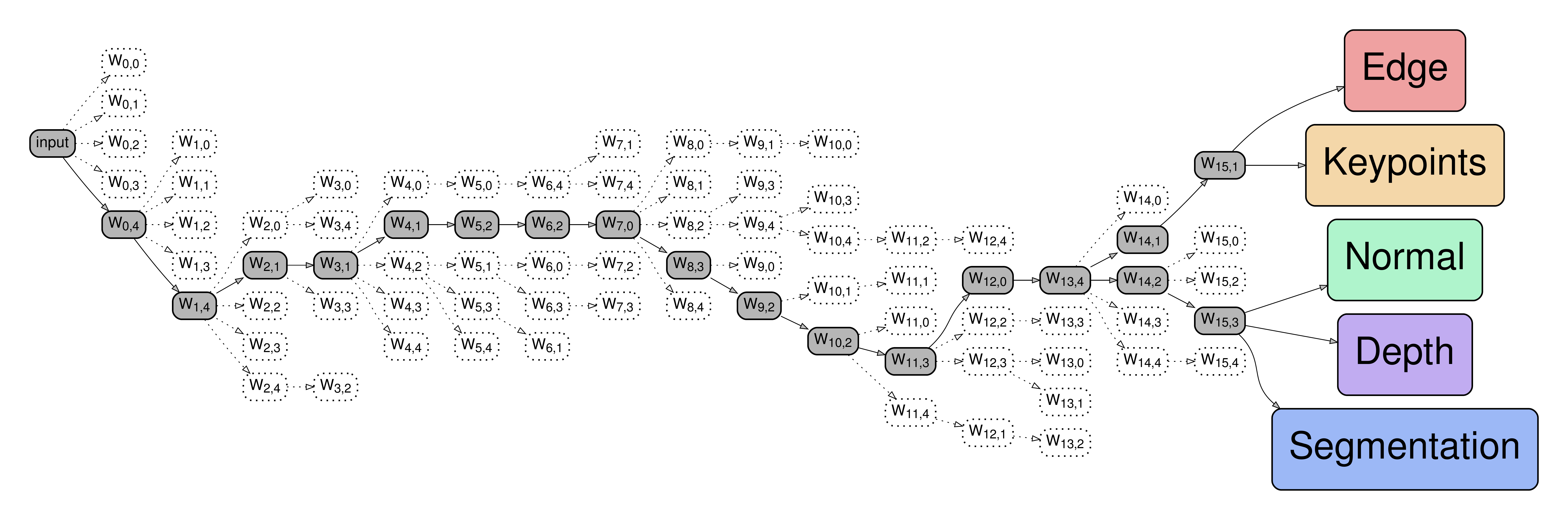}
	\vspace{-9mm}
    	\caption{}
	\label{taskonomy_4}
  \end{subfigure}
    \vspace{-3mm}
  \caption{Four randomly sampled network architectures trained on Taskonomy dataset. Our method discovers the same task grouping strategy in network (a), (c), and (d). Network (b) branches out at one layer later compared to the others but still shares the same task grouping strategy.}
  \label{taskonomy_archs}
    \vspace{-5mm}
\end{figure*}

\begin{figure*}[ht!]
\begin{center}
\vspace{2mm}
\centerline{\includegraphics[width=0.85\textwidth]{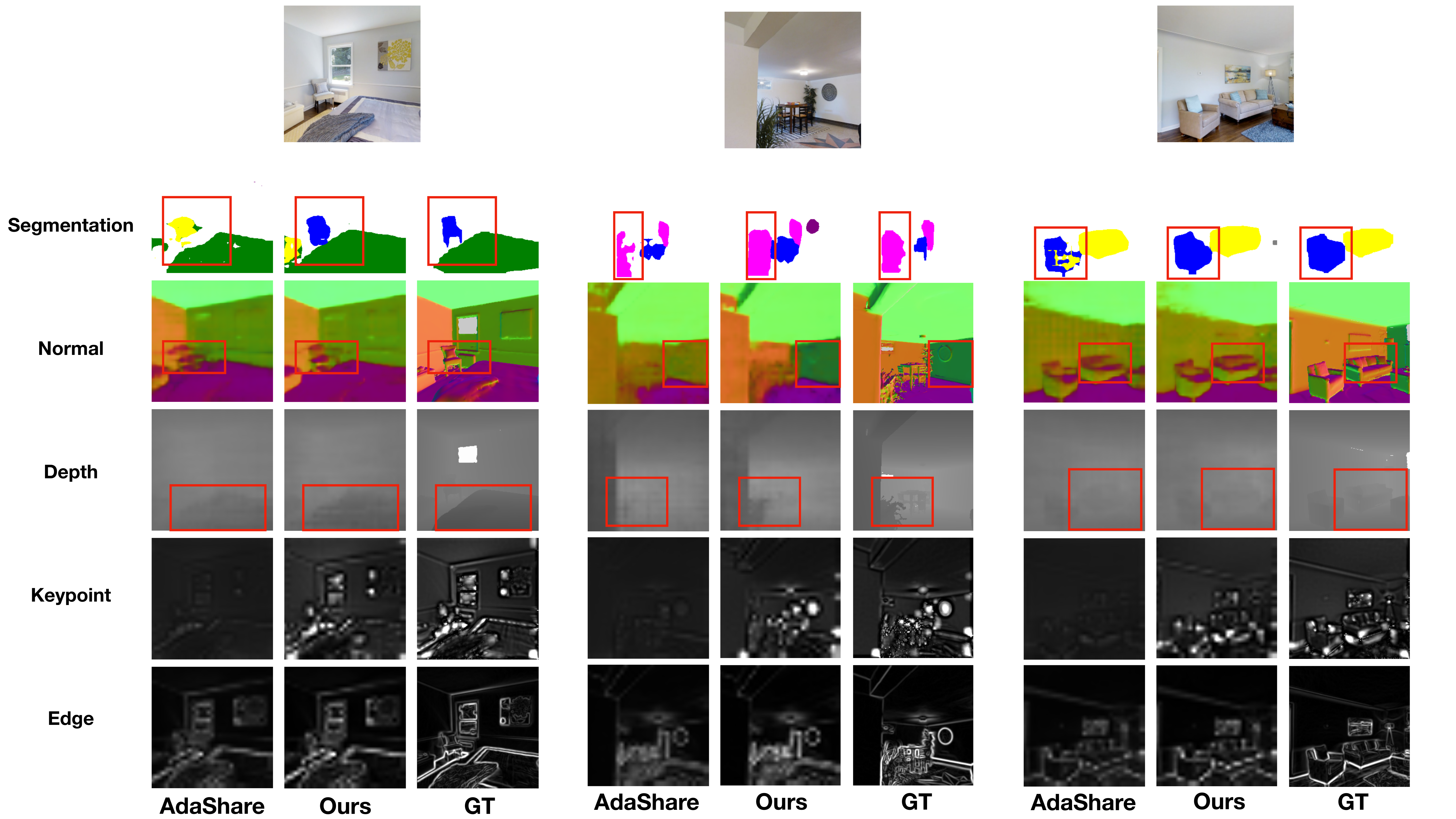}}
\vspace{-4mm}
\caption{Qualitative results of AdaShare~\cite{sun2019adashare}, our method, and ground truth. Our multi-task apporach produces cleaner predictions for high-level tasks (segmentation, normal, depth) and more accurate confidence scores for low-level tasks (keypoint, edge). }
\vspace{-9mm}
\label{taskonomy_visualization}
\end{center}
\end{figure*}

\textbf{Implementation details.}
We follow~\cite{sun2019adashare} to use the ResNet-34~\cite{he2016deep} backbone and the ASPP decoder~\cite{chen2017deeplab} for task-specific dense predictions. 
We use L1 loss for edge detection, keypoint detection and monocular depth tasks, cross-entropy loss for semantic segmentation task, and cosine similarity loss for surface normal  task. 

During the topology searching, we allow the network to branch at the end of every ResNet block. Each branching block has $5$ child nodes so the network capacity is similar to the baseline model in~\cite{sun2019adashare}.
We use the Adam optimizers with mini-batch size $64$ to train our network. Temperature is set to $10$ and decayed by the number of epochs. We again warmup the training for 2 epochs without updating the branching probabilities.
We perform grid search for learning rates in $(10^{-4}, 10^{-3}, 10^{-2})$ for both the weights and branching distributions, and weight decay in $(10^{-5}, 10^{-4}, 10^{-3})$. 
We train the network topology distribution on input image size of $128 \times 128$ and re-train the final selected network on image size of $256 \times 256$ for comparisons on the same image resolution. The input is normalized in $[-1, 1]$ and augmented by random clipping, scaling, and cropping.
On a single $32$GB Tesla GPU, it takes $2$ days to train the topology distribution and $3$ days to obtain the final converged network.

\textbf{Results.}
Following~\cite{chen2017deeplab, standley2019tasks, sun2019adashare}, we use the cross-entropy metrics with uncertain and background pixels masked out for segmentation; we use cosine similarities between the predictions and ground truth vectors without any masks for surface normal; we calculate the absolute mean error between the output and normalized ground truth with pixels whose depths are more than 126m masked out for depth; for the rest of the tasks, we calculate the absolute mean error between the output and normalized ground truth without any mask.

Table~\ref{taskonomy-results} lists the results that are based on the same evaluation protocol. We can clearly see that our method achieves the best performance on all $5$ tasks compared to all recent baselines. 
The first row in Table~\ref{taskonomy-results} shows the Single-Task setting where each task has its own set of parameters. It achieves lower performance than ours while having more than double parameter count. Our method is also more efficient than the recent Cross-Stitch and MTAN benchmarks as shown in the Table. AdaShare~\cite{sun2019adashare} uses a single-path network with adaptive skip-connections for multi-task learning. While the method has slightly less number of parameters, it achieves lower performance than our method on all $5$ tasks, showing the importance of feature sharing and branching.

We randomly sample four converged architectures after training and visualize them in Figure~\ref{taskonomy_archs}. 
We observe that even though the data flows take different paths in the four architectures, the final pruned network topology remains very similar to each other. 
Figure \ref{taskonomy_1}, \ref{taskonomy_3}, \ref{taskonomy_4} share exactly the same tree structure with edge and keypoint branch out from normal, depth and segmentation at the second last layer. 
Figure \ref{taskonomy_2} has a slightly different configuration that the branching occurs at the last layer. However, all the searched architectures show similar strategies for task grouping that are automatically found by the proposed method.

To further validate the effectiveness of the proposed branching operation, we directly sample a new network architecture without training the branching probability using the same topological space. Figure \ref{taskonomy_visualization} shows the qualitative results of the output from AdaShare~\cite{sun2019adashare}, the output from our method, and the ground truth.
We can see that our converged network produces better results visually, especially for segmentation, depth estimation, keypoint prediction and edge detection tasks.

\vspace{-1mm}
\section{Conclusion}
\vspace{-1mm}
In this work, we introduce an automated multi-task learning framework that learns the underlying task grouping strategies by sharing and branching a neural network. We propose a carefully designed topological space to enable direct optimization for both the weights and branching distributions of the network through gumbel-softmax sampling.
We validate the proposed method on controlled synthetic data, real-world CelebA, and large-scale Taskonomy dataset.
Future work includes extension of our approach to multi-modality inputs and tasks with partial annotations.

\bibliography{example_paper}
\bibliographystyle{icml2020}

\clearpage
\begin{appendix}
\section*{Appendix}
\section{Implementation Details}
In this section, we provide additional implementation details for experiments on CelebA dataset~\cite{liu2015faceattributes} and Taskonomy dataset~\cite{zamir2018taskonomy}.

\vspace{2mm}
\textbf{CelebA.}\\
(a) LearnToBranch-VGG network: we train the network topological distribution for $30$ epochs. The global learning rate is set to $10^{-5}$ and the learning rate for branching operation is set to $10^{-4}$. We use exponential learning decay with decay factor $0.97$ for every $2.4$ epochs.
After sampling the final architecture, we train the network for $30$ epochs from scratch. We set the learning rate to $0.03$, the weight decay to $5e^{-4}$, and the momentum to $0.9$. We decay the learning rate by half for every $10$ epoch.

(b) LearnToBranch-Deep-Wide network: we train the network topological distribution for $30$ epochs. The global learning rate is set to $10^{-4}$ and the learning rate for branching operation is set to $10^{-2}$. We use exponential learning decay with decay factor $0.97$ for every $2.4$ epochs.
After sampling the final architecture, we train the network for $30$ epochs from scratch. We set the learning rate to $0.05$, the weight decay to $5e^{-4}$, and the momentum to $0.9$. We decay the learning rate by half for every $15$ epoch.

\begin{figure*}
    \centering
  \begin{subfigure}[t]{.50\textwidth}
    	\centering\includegraphics[width=\textwidth]{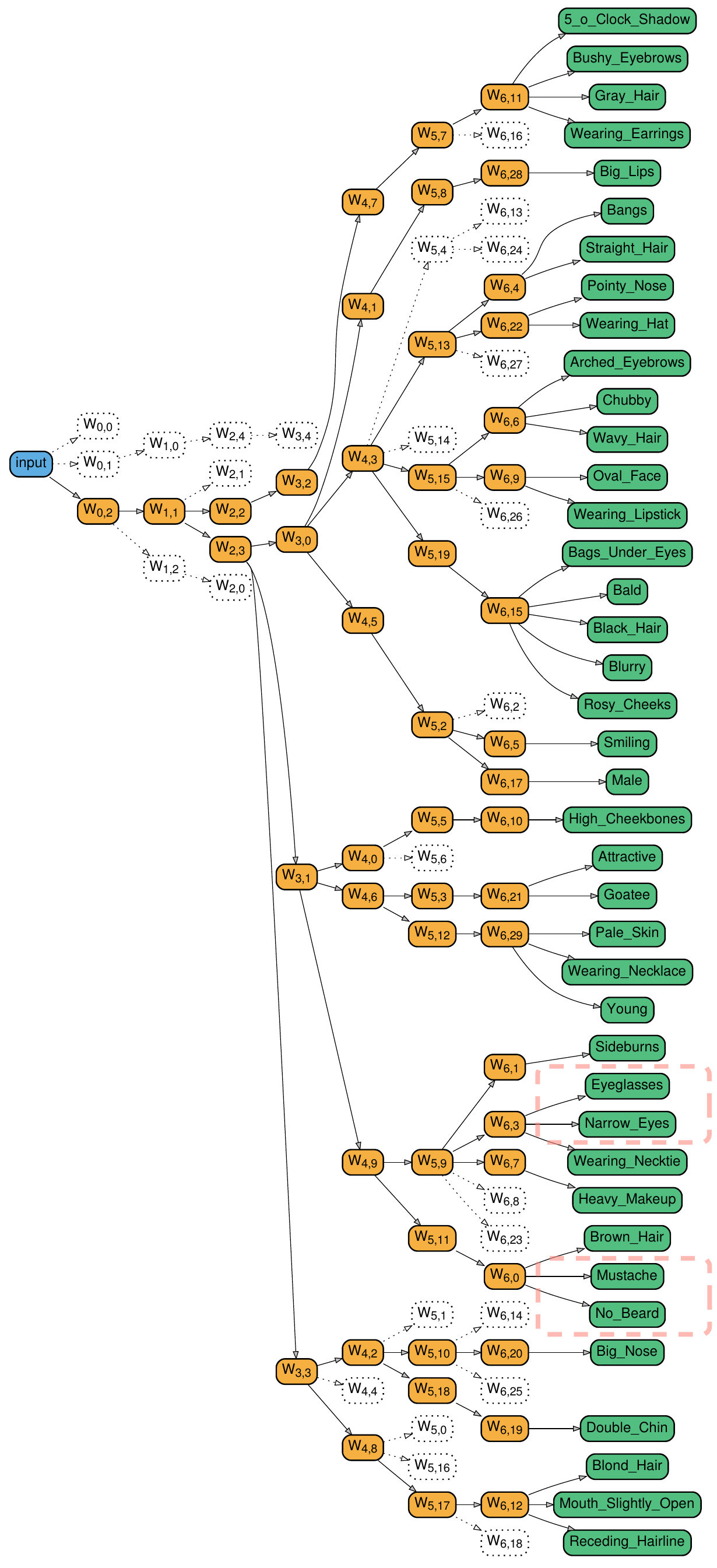}
   	 \caption{LearnToBranch-VGG}
	\label{thin32}
  \end{subfigure}
  \hspace{10mm}
  \begin{subfigure}[t]{.38\textwidth}
      	\centering\includegraphics[width=\textwidth]{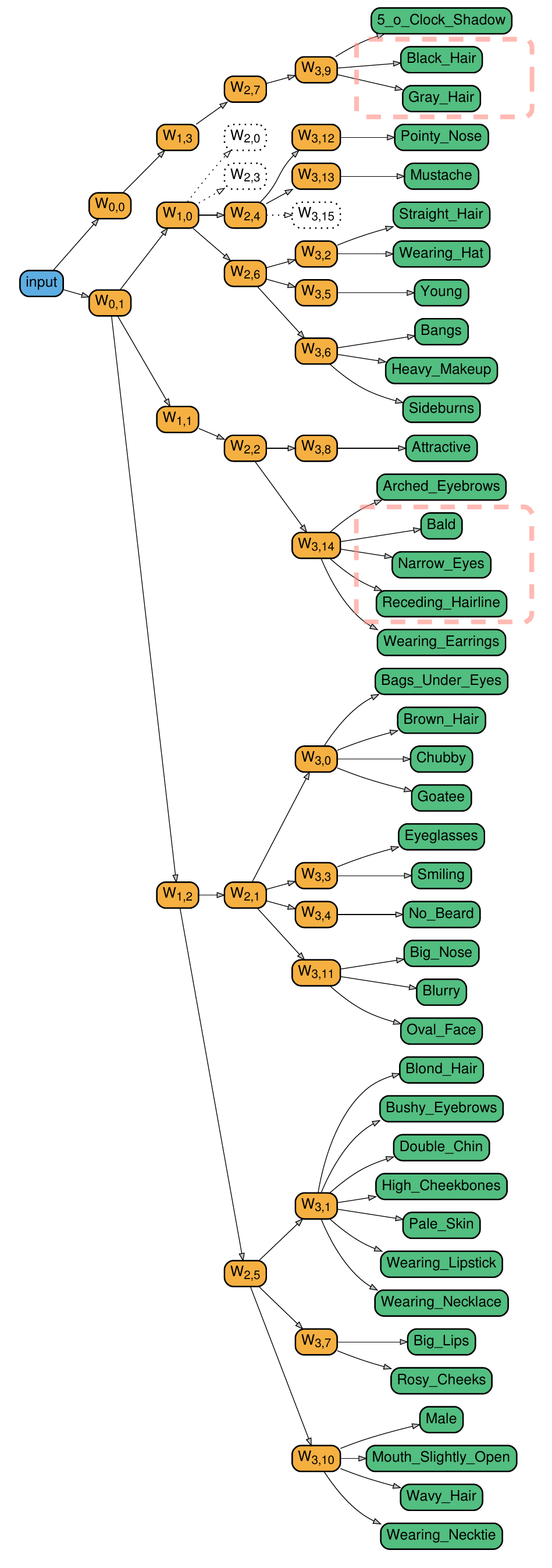}
    	\caption{LearnToBranch-Deep-Wide}
	\label{gnas_deep}
  \end{subfigure}
  \caption{Network architectures learned from CelebA dataset. We observe some grouping strategies learned by our method share some similarities with human intuition. For instance, network (a) groups 'Eyeglasses' and 'Narrow Eyes' and groups 'Mustasche' and 'No Beard'. Network (b) groups 'Black Hair' and 'Gray Hair' and groups 'Bald' and 'Receding Hairline'. The groups are shown in red dotted rectangles. Transparent boxes denote removed nodes because they are not selected by any child nodes.}
 	\label{celeba-archs}
\end{figure*}

We visualize both network architectures (a) and (b) in Figure \ref{celeba-archs}. 
We observe some grouping strategies learned by our method share some similarities with human intuition. For instance, network (a) groups 'Eyeglasses' and 'Narrow Eyes' and groups 'Mustasche' and 'No Beard'. Network (b) groups 'Black Hair' and 'Gray Hair' and groups 'Bald' and 'Receding Hairline'.

\vspace{2mm}
\textbf{Taskonomy.}\\
We train the network topological distribution for $30$ epochs. The global learning rate is set to $10^{-3}$, the learning rate for branching operations is set to $10^{-1}$, and weight decay is set to $10^{-5}$. We use exponential learning decay with decay factor $0.97$ for every $1$ epoch.

After sampling the final architecture, we train the network for $30$ epochs from scratch. We set the learning rate to $5e^{-4}$, the weight decay to $10^{-4}$, and the momentum to $0.9$. We use exponential learning decay with decay factor $0.97$ for every $1$ epoch.

We follow the work in~\cite{sun2019adashare} and set the following task weightings: $1.0$ for semantic segmentation, $3.0$ for surface normal estimation, $2.0$ for depth estimation, $7.0$ for keypoint prediction, and $7.0$ for edge detection. Note that we can further combine the proposed method with other adaptive task weighting methods. We leave this effort for future investigation.

Again following~\cite{sun2019adashare},
for the semantic segmentation task, we ignore uncertain pixels (class $0$) and background pixels (class $1$).
For the monocular depth estimation task, we ignore pixels with depth value larger than $64500$ and normalize the disparities by taking the $\log$ operation and downscale by a factor of $\log(2^{16})$.
For the surface normal prediction task, we normalize the three-dimensional normal vector from $[0, 255]$ to $[-1, 1]$.
For the keypoint estimation and the edge detection tasks, we downscale the original values by a factor of $2^{16}$. We then normalize the values from $[0, 0.005]$ to $[-1, 1]$ for keypoints and from $[0, 0.08]$ to $[-1, 1]$ for edges.

\section{Learned Branching Features}
We use Network Dissection~\cite{netdissect2017} to examine the features learned from Taskonomy dataset. We found that the SDN \{segmentation, depth, normal\} branch shows $35\%$ increase in high-level features (object and part detectors) and $20\%$ decrease in low-level features (texture detectors) compared to the shared layer before splitting. On the other hand, the EK \{edge, keypoint\} branch continues to focus on low-level features, showing no increase in high-level features due to the fact that \{edge, keypoint\} tasks are generally considered low-level tasks. Table~\ref{netdissect-results} lists the number of detector counts before and after the branching (layer $13$).
\vspace{-5mm}

\begin{table}[h]
\caption{Detector counts for different categories of input images at different layers using Network Dissection~\cite{netdissect2017}.}
\label{netdissect-results}
\begin{center}
\begin{scriptsize}
\begin{sc}
\begin{tabular}{lcccr}
\toprule
Layer & Object+Part Detectors & Texture Detectors \\
\midrule
$\text{layer}_{13}$ & 116 & 262\\
$\text{layer}_\text{14, SDN}$ & 157 & 208\\
$\text{layer}_\text{14, EK}$ & 118 & 253\\
\bottomrule
\end{tabular}
\end{sc}
\end{scriptsize}
\end{center}
\vskip -0.1in
\end{table}

\section{Generalizability of the Learned Branching}
We investigate whether the task grouping strategy learned from Tasknomoy dataset can be transferred to NYUv2 dataset on the three shared tasks across the two datasets. Following the metrics in Table 2, for \{segmentation, normal, depth\} tasks, we found that the grouping learned from Tasknomoy achieves \{1.611, 0.739, 0.058\} on NYUv2 test set while the grouping learned from NYUv2 training set achieves \{1.572, 0.748, 0.058\} on NYUv2 test set. The overall performance difference is relatively small at $1.23\%$. The experiment is performed on the NYUv2 labelled dataset with 795 training images and 654 test images using $256\times256$ image resolution.

\end{appendix}

\end{document}